\documentclass[10pt,twocolumn,letterpaper]{article}

\usepackage{cvpr}
\usepackage{times}
\usepackage{epsfig}
\usepackage{graphicx}
\usepackage{amsmath}
\usepackage{amssymb}
\usepackage{array,colortbl,multirow,multicol,booktabs,ctable}
\DeclareMathOperator*{\argmin}{arg\,min}
\usepackage{gensymb}
\newcommand{\ignore}[1]{}

\usepackage[pagebackref=true,breaklinks=true,letterpaper=true,colorlinks,bookmarks=false]{hyperref}

 \cvprfinalcopy 

\def\cvprPaperID{8279} 

\ifcvprfinal\fi
\begin{document}


\title{A simple baseline for domain adaptation using rotation prediction}

\author{Ajinkya Tejankar \qquad  Hamed Pirsiavash\\
University of Maryland Baltimore County\\
{\tt\small \{at6, hpirsiav\}@umbc.edu}
}

\maketitle

\begin{abstract}


Recently, domain adaptation has become a hot research area with lots of applications. The goal is to adapt a model trained in one domain to another domain with scarce annotated data. We propose a simple yet effective method based on self-supervised learning that outperforms or is on par with most state-of-the-art algorithms, e.g. adversarial domain adaptation. Our method involves two phases: predicting random rotations (self-supervised) on the target domain along with correct labels for the source domain (supervised), and then using self-distillation on the target domain. Our simple method achieves state-of-the-art results on semi-supervised domain adaptation on DomainNet dataset.

Further, we observe that the unlabeled target datasets of popular domain adaptation benchmarks do not contain any categories apart from testing categories. We believe this introduces a bias that does not exist in many real applications. We show that removing this bias from the unlabeled data results in a large drop in performance of state-of-the-art methods, while our simple method is relatively robust.


\end{abstract}

\section{Introduction}

\begin{figure}[t]
\centering
\includegraphics[height=.8 \textwidth]{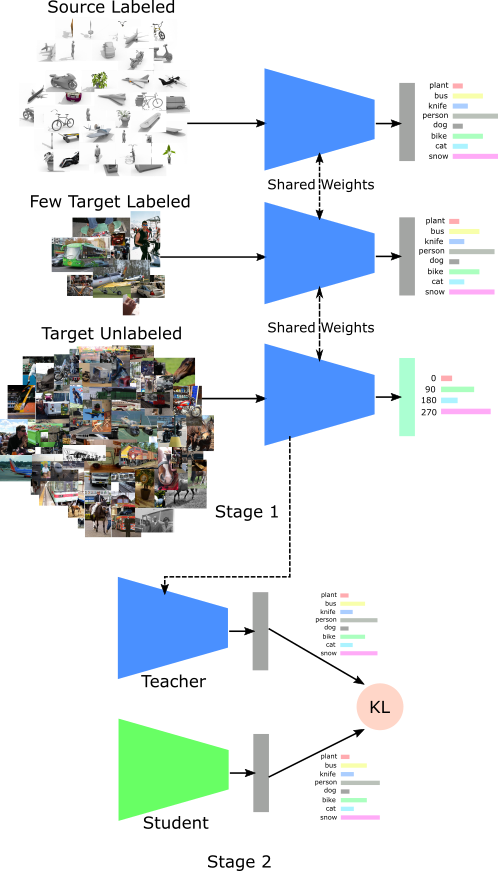}
\caption{Illustration of our proposed method for semi-supervised domain adaptation setting. We do supervised learning and rotation prediction in the first stage and then do knowledge distillation in the second stage.}
\vspace{-.2in}
\label{fig:teaser}
\end{figure}

\begin{figure*}[t]
\centering
\includegraphics[height=.55 \textwidth]{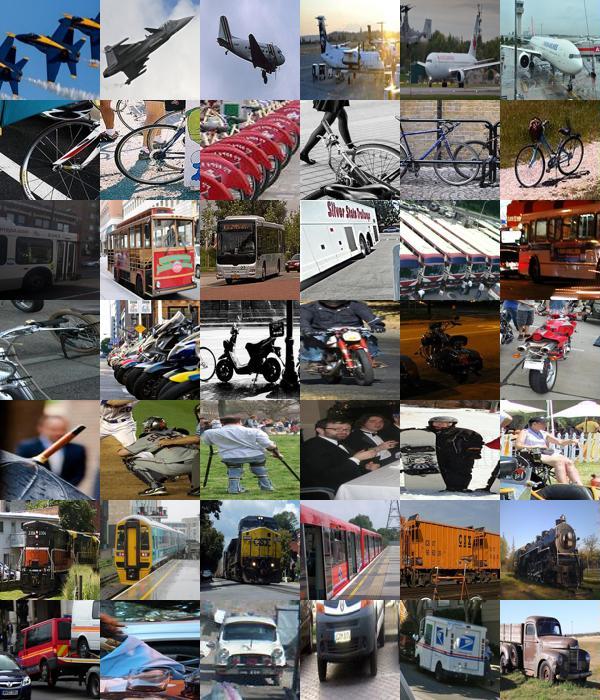}
\includegraphics[height=.55 \textwidth]{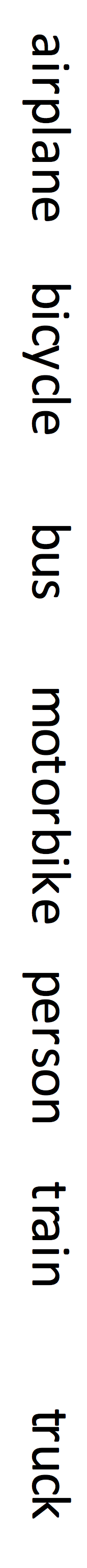}
\includegraphics[height=.55 \textwidth]{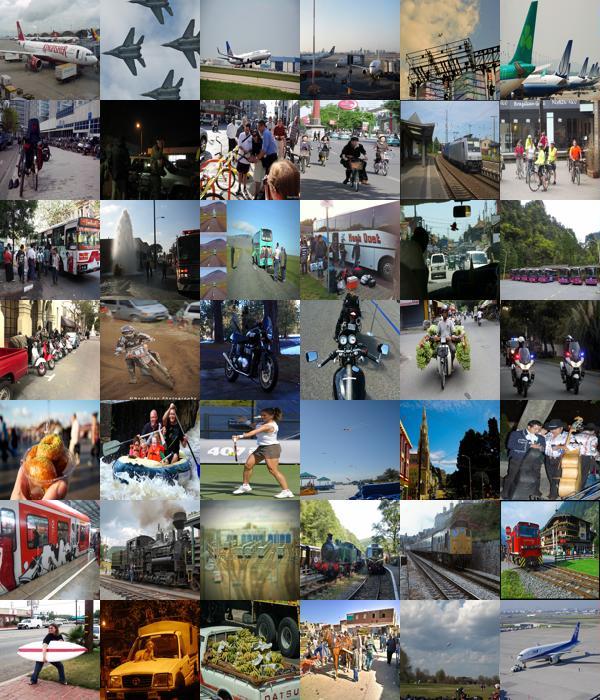}
\caption{Curated and un-curated unlabeled data for VisDa-17 dataset. {\bf left:} Random samples from the unlabeled target domain dataset used in VisDa-17 benchmark. They are originally sampled from MS-COCO dataset and then the objects of interest are cropped and centered. {\bf right:} Random samples for the same categories from MS-COCO dataset without cropping. We believe this cropping process injects bias into the unlabeled dataset that does not exist in practical applications. Such a bias in the benchmark can produce misleading conclusions as some methods mat exploit this bias. In our un-curated experiments, we not only use un-cropped images but also use images from other unknown categories.}
\label{patch}
\end{figure*}

Consider the following problem: we have an image classification model trained on synthetic images (we call source domain) and we want to use this model to classify real images (we call target domain) to the same categories. The model has never seen the real images, so naturally, it does not perform well on real images as the model is highly tuned to synthetic images. This difference in performance between source and target domain is called domain gap and the technique of adapting the classifier to overcome this domain gap is called domain adaptation (DA). 

In domain adaptation, we assume the training has access to a large scale unlabeled dataset from the target domain and just a small or even no annotated dataset from the target domain. The goal is to adapt the model to close the domain gap by performing well on the target domain. In general, it is assumed that the set of categories in the target domain are the same as those in the source domain. In this work, we mainly focus on the task of domain adaptation for image classification.\\

\noindent {\bf Motivation:}

\noindent Recently, unsupervised domain adaptation (UDA) has become a hot research topic due to its various applications in real world where it is difficult and costly to annotate images in specific application domains. For instance, assume a customer buys a household robot for which the visual perception system is trained at the factory. However, the appearance of objects in the customer's house may be different from the training data due to lighting and other instance specific variations. This will result in degraded robot vision. Hence, one can improve the robot's visual perception by collecting some unlabeled data from the customer's house and then adapting the model to this new domain using domain adaptation algorithms.

As another example, reinforcement learning (RL) has recently shown a lot of promise in various applications. However, most RL methods require lots of trials which are not possible in the real world due to physical limitations. Hence, most RL methods are being trained on graphics simulators and then tested in the real world. Clearly, the difference between synthetic and real data will lead to a domain gap in this setting. Consequently, domain adaptation can be used to reduce the domain gap.


\noindent {\bf Self-supervised learning:}

\noindent There is a general class of methods called self-supervised learning that focuses on learning rich representation exclusively from unlabeled data. The idea is to use inherent biases in natural images to design a pseudo task for which the annotation is automatic. Then, by forcing the model to solve the pseudo task, the model will learn rich features that capture the manifold of natural images. For instance, since most natural images are upright, we, humans, can detect if an image is rotated, so in RotNet \cite{gidaris2018unsupervised}, we can randomly rotate the images and train a model to predict the applied rotation angle. This pseudo-task does not need any annotations and learns rich visual features.

Most works in self-supervised learning evaluate the representations learned at each layer of the network by training a linear classifier, e.g., ImageNet classification. Usually, the fully supervised model performs better than the self-supervised model, but the gap shrinks as we go to the earlier layers. This is expected as the late layers are specifically tuned for the pseudo-task rather than general visual recognition. For instance in Table 3 of \cite{noroozi2017representation}, the first layer of the self-supervised model performs even slightly better than the fully supervised model on the linear classification task. 

\noindent {\bf Our intuition:}

\noindent Our intuition is that in adapting a model trained on the source domain to solve the same task in the target domain, tuning early layers is more important as the final layers are connected to the semantics and do not need to modify much. For instance, in ``car'' detection, the early layers detect intermediate concepts like ``wheels'', ``head-lights'', and ``windshield'' and the late layers compose those concepts together to come up with ``car'' detection. Hence, in domain adaptation, detectors of those intermediate concepts (early layers) need to be modified while the composition (late layers) may still be valid. This is more obvious to think about when we use real images for the source domain and clipart images for the target domain.

Therefore, we propose a simple baseline for domain adaptation where we train a model on the source domain and then adapt it to the target domain by solving a self-supervised learning method, namely rotation prediction, on the target domain. Since we do not want to forget high-level knowledge learned in the source domain, we keep learning the supervised task in the source domain in a multi-task learning setting.

In this work, we show that this simple method can be a very competitive baseline that either outperforms or is on par with most state-of-the-art methods on two popular benchmarks, (DomainNet \cite{Peng_2019_ICCV} and VisDa-17 \cite{Peng2017VisDATV}). Note that most state-of-the-art method on DA employ complicated algorithms, adversarial learning, that are not straight forward for optimization and reproducibility. Therefore, we believe our simple baseline can have a positive impact on the design of future DA methods.

Moreover, we observe that, interestingly, most DA benchmarks have a selection bias in their unlabeled target domain that may not be present in real world applications. This bias can be exploited by DA methods resulting in misleading benchmarking. Our empirical study shows that removing this bias, which is easily possible, leads to degraded performance for most methods while our simple baseline is relatively more robust. We will discuss this bias more in the following section.

\section{Un-curated unlabeled data}
We believe using unlabeled data for domain adaptation is a practical setting that can be used in many applications where collection of unlabeled data in the target domain is easy and almost for free. In such settings, there is no effort required for data annotation, so the data can be from any category. For instance, we train a classifier on the synthetic data for $n$ categories and want to adapt it to the real data to perform classification for those $n$ categories. However, in practice the unlabeled real data may come from any category. Even outside the $n$ categories of interest.

Interestingly, popular DA benchmarks in the computer vision community create their unlabeled data by choosing images of the $n$ categories and then removing the labels. This strategy is used in DomainNet \cite{Peng_2019_ICCV} which is one of the most recent benchmarks and also in VisDa-17 \cite{Peng2017VisDATV} that is also a well-known benchmark. Even more interestingly, in VisDa-17 dataset, not only the unlabeled data (adopted from MS-COCO dataset) is from the same $n$ categories of interest, but also the images are cropped to contain only the bounding box of those objects of interest. Note that the same strategy is used in most self-supervised learning literature when using ImageNet dataset with no labels as the source of unlabeled data. We call this a curated (standard) unlabeled data and we call an unlabeled dataset that may contain any object category as an un-curated data. Fig. \ref{patch} shows some samples from standard and our un-curated VisDa-17 dataset for comparison.

We believe, using curated unlabeled data is not a good idea and is not aligned with the final practical applications. Hence, it can be misleading as a benchmark. The problem is that the process of curation can be seen as a form of weak supervision leaked into the unlabeled training data that may not exist in the real application. Since all algorithms will not exploit this weak supervision, the resulting benchmark can be misleading.

For instance, the entropy minimization method \cite{grandvalet2005semi} chooses the entropy of the output as the loss to encourage the model to produce a prediction with low uncertainty. This is great in the case of curated data as we know that each unlabeled image corresponds to one of the known categories. However, in the case of un-curated data, the model may be uncertain for data from other unknown categories. Thus, minimizing the entropy loss may not be a good idea. We support this hypothesis in our experiments by adding unknown categories from the target domain to the standard unlabeled data.

\section{Related Work}

\noindent \textbf{Domain adaptation:}

\noindent A popular strategy for dealing with the domain gap is to learn features that are consistent across domains. One of the most popular methods of aligning features for both domains is adversarial training of a discriminator and a feature extractor such that the discriminator cannot distinguish between the features of source and target domains \cite{chen2019transferability,Pei2018MultiAdversarialDA,Tzeng2017AdversarialDD,Kumar2018CoregularizedAF, Long2017ConditionalAD, saito2018adversarial,Pinheiro2017UnsupervisedDA,Lee_2019_ICCV}. Our method is largely orthogonal to the above domain adaptation methods.

Since most of the domain adaptation work focuses on the unsupervised setting, semi-supervised domain adaptation (SSDA) is not well studied. In \cite{Saito_2019_ICCV}, the standard unsupervised domain adaption (UDA) methods were shown to be less effective in the SSDA setting. \cite{Saito_2019_ICCV} introduced an iterative algorithm that alternates between minimizing and maximizing the entropy of the output. Our method is different and simpler compared to \cite{Saito_2019_ICCV}. We show that our method does not degrade in performance when going from unsupervised to semi-supervised setting, achieving state-of-the-art results for SSDA on DomainNet, introduced in \cite{Saito_2019_ICCV}. 

Some other works have employed semi-supervised learning. In \cite{french2018selfensembling}, a network is trained to match the ensembled predictions of its own output obtained at different time intervals during training. Further, combination of adversarial training with semi-supervised techniques like entropy minimization \cite{grandvalet2005semi} and VAT \cite{Miyato2017VirtualAT} have been explored in \cite{shu2018a, Lee_2019_ICCV}.

\noindent \textbf{Self-supervised learning:}

\noindent Numerous pretext tasks, also called pseudo tasks, have been developed for unsupervised representation learning. \cite{Caron_2019_ICCV, gidaris2018unsupervised} predict image transformations. In \cite{Doersch_2015_ICCV,noroozi2016unsupervised}, spatial structure of the image is exploited to create pretext tasks. In \cite{noroozi2017representation}, a model is trained by enforcing count consistency in image and its tiles. In \cite{caron2018deep}, a model is iteratively trained to classify images based on the labels obtained using k-means clustering. In \cite{Noroozi_2018_CVPR}, a teacher network is trained on a hard pretext task and its knowledge is transferred to a student network via k-means clustering. For simplicity, we briefly evaluate the Jigsaw \cite{noroozi2016unsupervised} pretext task, but focus on the simpler, more effective pretext task of predicting rotations \cite{gidaris2018unsupervised}.

Aside from representation learning, auxiliary pretext-tasks can also help the model generalize better. In \cite{hendrycks2019using}, it is shown that incorporating self-supervised losses during pre-training can improve model robustness. In \cite{Chen_2019_CVPR}, rotation prediction is shown to effectively replace labeled data in conditional GANs to make them completely unsupervised. In \cite{Gidaris_2019_ICCV}, self-supervision is combined with existing few-shot learning methods to leverage the unlabeled data and boost the model performance. In \cite{Zhai_2019_ICCV}, self-supervision is applied to semi-supervised learning by incorporating a supervised loss on a small amount of labeled data while solving the pretext task on the entire dataset. Our simple method is similar to \cite{Zhai_2019_ICCV} except that we are applying it to the task of domain adaptation and improve it with self-distillation.


In \cite{Carlucci_2019_CVPR}, a modified implementation of the Jigsaw pretext task is used as an auxiliary task for domain generalization from multiple source domains to any target domain. We are interested in domain adaptation from a single source to a single target domain. In the experiments, we show that replacing our RotNet pseudo-task with Jigsaw does not perform well in our setting.


\noindent \textbf{Knowledge distillation:}

\noindent Knowledge distillation was originally proposed by \cite{caruna2006model}, and used in \cite{hinton2015distilling} to transfer the knowledge from one or more teacher networks to a single student network. \cite{bagherinezhad2018label,Furlanello2018BornAgainNN,2017AGF} show that self-distillation, where the teacher and student share the same architecture, improves supervised learning by reducing the generalization gap. We use this method to improve our simple baseline method for SSDA and UDA.

\section{Method}


We are interested in domain adaptation where we have lots of annotated data in the source domain, either small or no annotated data in the target domain, and large scale unlabeled data in the target domain. This is a practical assumption in many applications as unlabeled data is usually available abundantly. The goal is to train a model on the source domain and then adapt it to the target domain using both labeled and unlabeled data. Recently, this problem has become popular due to its various practical applications e.g. in learning from synthetic data. 

We are proposing a simple baseline for this task using standard self-supervised learning algorithms. We train a network with two heads (one for image classification and one for rotation prediction) in a multi-task setting with the following three tasks:
(1) supervised learning on the source domain, (2) supervised learning on the target domain using a small annotated dataset, and (3) self-supervised learning on the target domain by predicting rotation applied to an unlabeled image. Fig. \ref{fig:teaser} shows our method for SSDA.

For the third task above, following RotNet method \cite{gidaris2018unsupervised}, given an unlabeled image, we rotate it using a rotation angle randomly chosen from the list $\{0^{\circ}, 90^{\circ}, 180^{\circ}, 270^{\circ}\}$, and then define its corresponding label to be the rotated angle (1 out of 4 possibilities). Then, we input rotated images to the network and optimize it to detect the rotation angle using the cross entropy loss function.

Finally, we perform standard knowledge distillation on the finetuned model using the unlabeled target data to improve the model by taking advantage of the soft pseudo-labels. We use the same architecture for both student and teacher networks. This is similar to \cite{bagherinezhad2018label,Furlanello2018BornAgainNN} except that we do distillation on unlabeled target data rather than labeled data. Similar to \cite{bagherinezhad2018label, Furlanello2018BornAgainNN}, our intuition is that this method will reduce the generalization gap caused by using one-hot encoding of the ground truth in the first stage.


More formally, given an image $x^s$ and its label $y^s$ in the source domain, an image $x^t$ and its label $y^t$ in the target domain, and also an unlabeled image $x^{u}$ in the target domain, we define the following loss terms:
\begin{align} \nonumber
\mathcal{L}^s_{sup}(f) &= \sum_i \ell_{ce} (f(x^s_i), y^s_i)\\ \nonumber
\mathcal{L}^t_{sup}(f) &= \sum_i \ell_{ce} (f(x^t_i), y^t_i)\\ \nonumber
\mathcal{L}^t_{ssl}(r) &= \sum_i \ell_{ce} (r(T_{a}(x^u_i)), a) \nonumber
\end{align}
\vspace{.05in}
    
\noindent where $\mathcal{L}^s_{sup}$ is the supervised loss on the source domain, $\mathcal{L}^t_{sup}$ is the supervised loss on the target domain,, $\mathcal{L}^t_{ssl}$ is the self-supervised loss on the target domain, $\ell_{ce}(.)$ is the cross entropy loss, $f(.)$ is the classifier, $r(.)$ is the rotation prediction classifier, and $T_a(.)$ is an operator that rotates the input image by an angle $a \in \{0^{\circ}, 90^{\circ}, 180^{\circ}, 270^{\circ}\}$ which is chosen randomly for each data point and iteration. Note that $f(.)$ and $r(.)$ share all layers except the last one.

In the case of SSDA, we optimize the following loss function:

\begin{align}
\argmin_{f,r} \Big(\lambda_{s} \mathcal{L}^s_{sup}(f) + \lambda_{t} \mathcal{L}^t_{sup}(f) + \lambda_{ssl} \mathcal{L}^t_{ssl}(r)\Big)
\label{opt}
\end{align}

Finally, we perform standard knowledge distillation by optimizing the following loss function:

$$ \argmin_g \sum_i \mathcal{L}_{ce} \big(KL(f(x^u_i) || g(x^u_i)\big)$$

\noindent where $g(.)$ is the student network and $f(.)$ is the teacher network that is learned in the first stage and frozen at knowledge distillation. Note that $f(.)$ and $g(.)$ share the same network architecture, but the weights in $g(.)$ are randomly initialized. Also, note that we can choose to distill from multiple teacher networks in an ensemble setting. We show some of such results in the appendix material. 

In the case of unsupervised domain adaptation, we remove $\mathcal{L}_{sup}^t$ term from Eq. \ref{opt}. 



{\bf Entropy minimization (ENT)} is proposed initially in \cite{grandvalet2005semi} as a form of semi-supervised learning. Assuming that the model should be confident about its prediction for the unlabeled data, we can minimize the entropy of the predicted probability on the unlabeled data. We add this loss term to our method and show that it helps the learning with a good margin on standard domain adaptation datasets. Hence, we add the following loss term to the optimization in Eq. \ref{opt}:

$$ \mathcal{L}^t_{ent}(f) =  \sum_i ent(f(x^u_i))$$


\noindent where $ent(p)$ calculates the entropy of distribution $p$.

\begin{table*}[t]
\begin{center}
\scalebox{0.85}
{
\renewcommand{\arraystretch}{1.1}
\begin{tabular}{l|cccccccccccccc|cc}
\toprule

\multirow{2}{*}{Method} &
\multicolumn{2}{c}{R to C} & \multicolumn{2}{c}{R to P} & \multicolumn{2}{c}{P to C} &
\multicolumn{2}{c}{C to S} & \multicolumn{2}{c}{S to P} & \multicolumn{2}{c}{R to S} &
\multicolumn{2}{c}{P to R} & \multicolumn{2}{|c}{Mean} \\

&
1\scriptsize{-shot} & 3\scriptsize{-shot} & 1\scriptsize{-shot} & 3\scriptsize{-shot} & 1\scriptsize{-shot} & 3\scriptsize{-shot} &
1\scriptsize{-shot} & 3\scriptsize{-shot} & 1\scriptsize{-shot} & 3\scriptsize{-shot} & 1\scriptsize{-shot} & 3\scriptsize{-shot} &
1\scriptsize{-shot} & 3\scriptsize{-shot} & 1\scriptsize{-shot} & 3\scriptsize{-shot}  \\
\hline

S+T & 43.3 & 47.1 & 42.4 & 45.0 & 40.1 & 44.9 & 33.6 & 36.4 & 35.7 & 38.4 & 29.1 & 33.3 & 55.8 & 58.7 & 40.0 & 43.4 \\
DANN \cite{Ganin:2015:UDA:3045118.3045244} & 43.3 & 46.1 & 41.6 & 43.8 & 39.1 & 41.0 & 35.9 & 36.5 & 36.9 & 38.9 & 32.5 & 33.4 & 53.6 & 57.3 & 40.4 & 42.4 \\
ADR \cite{saito2018adversarial} & 43.1 & 46.2 & 41.4 & 44.4 & 39.3 & 43.6 & 32.8 & 36.4 & 33.1 & 38.9 & 29.1 & 32.4 & 55.9 & 57.3 & 39.2 & 42.7 \\
CDAN \cite{Long2017ConditionalAD} & 46.3 & 46.8 & 45.7 & 45.0 & 38.3 & 42.3 & 27.5 & 29.5 & 30.2 & 33.7 & 28.8 & 31.3 & 56.7 & 58.7 & 39.1 & 41.0 \\
ENT \cite{grandvalet2005semi} & 37.0 & 45.5 & 35.6 & 42.6 & 26.8 & 40.4 & 18.9 & 31.1 & 15.1 & 29.6 & 18.0 & 29.6 & 52.2 & 60.0 & 29.1 & 39.8 \\
MME \cite{Saito_2019_ICCV} & 48.9 & \bf{55.6} & 48.0 & 49.0 & 46.7 & 51.7 & 36.3 & 39.4 & 39.4 & 43.0 & 33.3 & 37.9 & 56.8 & 60.7 & 44.2 & 48.2 \\
\hline
KD(ROT)     & 49.3 & 54.5 & 49.9 & \bf{51.3} & \bf{48.3} & 52.9 & \bf{39.7} & \bf{43.2} & \bf{44.3} & \bf{46.3} & \bf{40.0} & \bf{43.2} & 58.7 & 61.5 & \bf{47.2} & \bf{50.4} \\
KD(ROT+ENT) & \bf{51.0} & 54.7 & \bf{50.5} & 50.9 & 47.8 & \bf{53.0} & 37.7 & 42.3 & 38.1 & \bf{46.3} & 38.0 & 41.8 & \bf{60.4} & \bf{62.1} & 46.2 & 50.2 \\

\bottomrule
\end{tabular}
}
\end{center}
\caption{SSDA on standard DomainNet for AlexNet: S+T is the baseline that uses only supervised learning on both domains. The domain names: {\bf R:} Real, {\bf C:} Clipart, {\bf P:} Painting, {\bf S:} Sketch. We study 1-shot and 3-shot settings and compare our method to various baselines. Our method, even though simple, outperforms the state-of-the-art baselines.}
\label{tb:dn_curated_alexnet}

\end{table*}

\begin{table*}[t]
\begin{center}
\scalebox{0.85}
{
\renewcommand{\arraystretch}{1.2}
\begin{tabular}{l|cccccccccccccc|cc}
\toprule
\multirow{2}{*}{Method} &
\multicolumn{2}{c}{R to C} & \multicolumn{2}{c}{R to P} & \multicolumn{2}{c}{P to C} &
\multicolumn{2}{c}{C to S} & \multicolumn{2}{c}{S to P} & \multicolumn{2}{c}{R to S} &
\multicolumn{2}{c}{P to R} & \multicolumn{2}{|c}{Mean} \\

 &
1\scriptsize{-shot} & 3\scriptsize{-shot} & 1\scriptsize{-shot} & 3\scriptsize{-shot} & 1\scriptsize{-shot} & 3\scriptsize{-shot} &
1\scriptsize{-shot} & 3\scriptsize{-shot} & 1\scriptsize{-shot} & 3\scriptsize{-shot} & 1\scriptsize{-shot} & 3\scriptsize{-shot} &
1\scriptsize{-shot} & 3\scriptsize{-shot} & 1\scriptsize{-shot} & 3\scriptsize{-shot}  \\
\hline

S+T         & 55.6 & 60.0 & 60.6 & 62.2 & 56.8 & 59.4 & 50.8 & 55.0 & 56.0 & 59.5 & 46.3 & 50.1 & 71.8 & 73.9 & 56.9 & 60.0 \\
DANN \cite{Ganin:2015:UDA:3045118.3045244} & 58.2 & 59.8 & 61.4 & 62.8 & 56.3 & 59.6 & 52.8 & 55.4 & 57.4 & 59.9 & 52.2 & 54.9 & 70.3 & 72.2 & 58.4 & 60.7 \\
ADR \cite{saito2018adversarial} & 57.1 & 60.7 & 61.3 & 61.9 & 57.0 & 60.7 & 51.0 & 54.4 & 56.0 & 59.9 & 49.0 & 51.1 & 72.0 & 74.2 & 57.6 & 60.4 \\
CDAN \cite{Long2017ConditionalAD} & 65.0 & 69.0 & 64.9 & 67.3 & 63.7 & 68.4 & 53.1 & 57.8 & 63.4 & 65.3 & 54.5 & 59.0 & 73.2 & 78.5 & 62.5 & 66.5 \\
ENT \cite{grandvalet2005semi}        & 65.2 & 71.0 & 65.9 & 69.2 & 65.4 & 71.1 & 54.6 & 60.0 & 59.7 & 62.1 & 52.1 & 61.1 & 75.0 & 78.6 & 62.6 & 67.6 \\
MME \cite{Saito_2019_ICCV} & \bf{70.0} & \bf{72.2} & 67.7 & 69.7 & \bf{69.0} & \bf{71.7} & 56.3 & 61.8 & \bf{64.8} & 66.8 & \bf{61.0} & 61.9 & \bf{76.1} & 78.5 & \bf{66.4} & 68.9 \\
\hline
KD(ROT)     & 63.6 & 64.0 & 66.5 & 67.0 & 60.8 & 65.0 & 57.5 & 60.9 & 63.4 & 62.6 & 58.3 & 60.3 & 74.6 & 75.6 & 63.5 & 65.1 \\
KD(ROT+ENT) & 65.6 & 71.6 & \bf{70.4} & \bf{70.8} & 64.8 & 71.2 & \bf{58.1} & \bf{64.1} & 62.6 & \bf{67.4} & 60.0 & \bf{63.6} & 75.8 & \bf{80.7} & 65.3 & \bf{69.9} \\
\bottomrule

\end{tabular}
}
\end{center}
\caption{SSDA on standard DomainNet for ResNet34. The description is similar to Table \ref{tb:dn_curated_alexnet}. Our method (last row) outperforms all the baselines in average for 3-shot setting.}
\label{tb:dn_curated_resnet}

\end{table*}

\begin{table*}[t]
\begin{center}
\scalebox{0.85}
{
\renewcommand{\arraystretch}{1.2}
\begin{tabular}{lc|cccccccccccccc|cc}
\toprule

\multirow{2}{*}{Method} & \multirow{2}{*}{\hspace{-.2in}Data} &
\multicolumn{2}{c}{R to C} & \multicolumn{2}{c}{R to P} & \multicolumn{2}{c}{P to C} &
\multicolumn{2}{c}{C to S} & \multicolumn{2}{c}{S to P} & \multicolumn{2}{c}{R to S} &
\multicolumn{2}{c}{P to R} & \multicolumn{2}{|c}{Mean} \\

& &
1\scriptsize{-shot} & 3\scriptsize{-shot} & 1\scriptsize{-shot} & 3\scriptsize{-shot} & 1\scriptsize{-shot} & 3\scriptsize{-shot} &
1\scriptsize{-shot} & 3\scriptsize{-shot} & 1\scriptsize{-shot} & 3\scriptsize{-shot} & 1\scriptsize{-shot} & 3\scriptsize{-shot} &
1\scriptsize{-shot} & 3\scriptsize{-shot} & 1\scriptsize{-shot} & 3\scriptsize{-shot}  \\
\hline

ENT (rerun)   & S & 41.5 & 42.9 & 36.3 & 42.8 & 24.8 & 39.0 & 21.6 & 33.3 & 18.8 & 35.2 & 17.7 & 30.8 & 53.6 & 60.3 & 30.6 & 40.6 \\
MME (rerun)   & S & 43.7 & 45.2 & 45.4 & 47.2 & 43.8 & 47.9 & 36.8 & 40.9 & 40.6 & 45.2 & 33.8 & 36.9 & 56.9 & 60.4 & 43.0 & 46.2 \\
KD(ROT+ENT)   & S & 51.0 & 54.7 & 50.5 & 50.9 & 47.8 & 53.0 & 37.7 & 42.3 & 38.1 & 46.3 & 38.0 & 41.8 & 60.4 & 62.1 & 46.2 & 50.2 \\
KD(ROT)      & S & 49.3 & 54.5 & 49.9 & 51.3 & 48.3 & 52.9 & 39.7 & 43.2 & 44.3 & 46.3 & 40.0 & 43.2 & 58.7 & 61.5 & 47.2 & 50.4 \\
\hline\hline
ENT   & U & 30.6 & 40.5 & 33.8 & 40.9 & 25.2 & 33.3 & 22.2 & 30.7 & 16.7 & 31.6 & 17.4 & 26.9 & 50.8 & 57.5 & 28.1 & 37.3 \\
MME    & U & 41.2 & 45.8 & 43.0 & 44.3 & 39.4 & 43.5 & 34.1 & 37.1 & 39.7 & 42.0 & 30.9 & 34.7 & 53.4 & 57.8 & 40.3 & 43.6 \\
KD(ROT+ENT)   & U & 46.4 & 48.9 & 47.2 & \bf{50.0} & 43.7 & 49.0 & 34.3 & 39.2 & 35.8 & 41.8 & 33.3 & 38.1 & 57.9 & \bf{60.5} & 42.7 & 46.8 \\
KD(ROT)      & U & \bf{47.7} & \bf{51.3} & \bf{48.9} & 48.3 & \bf{46.5} & \bf{50.2} & \bf{37.4} & \bf{41.7} & \bf{42.7} & \bf{43.8} & \bf{38.1} & \bf{40.5} & \bf{58.5} & 60.3 & \bf{45.7} & \bf{48.0} \\

\bottomrule
\end{tabular}
}
\end{center}
\caption{SSDA on un-curated DomainNet for AlexNet. Please see Table \ref{tb:dn_curated_alexnet} for the description of names. For better comparison, we rerun the baselines using their publicly released code. The top section uses the standard data denoted by "S", and the second section uses un-curated data denoted by "U". In average, our method (KD(ROT)) outperforms the baselines on the un-curated data. As expected adding ENT degrades the performance on the un-curated data.}
\label{tb:dn_uncurated_alexnet}

\end{table*}

\begin{table*}[t]
\begin{center}
\scalebox{0.85}
{
\renewcommand{\arraystretch}{1.2}
\begin{tabular}{lc|cccccccccccccc|cc}
\toprule
\multirow{2}{*}{Method} & \multirow{2}{*}{\hspace{-.2in}Data} &
\multicolumn{2}{c}{R to C} & \multicolumn{2}{c}{R to P} & \multicolumn{2}{c}{P to C} &
\multicolumn{2}{c}{C to S} & \multicolumn{2}{c}{S to P} & \multicolumn{2}{c}{R to S} &
\multicolumn{2}{c}{P to R} & \multicolumn{2}{|c}{Mean} \\

 & &
1\scriptsize{-shot} & 3\scriptsize{-shot} & 1\scriptsize{-shot} & 3\scriptsize{-shot} & 1\scriptsize{-shot} & 3\scriptsize{-shot} &
1\scriptsize{-shot} & 3\scriptsize{-shot} & 1\scriptsize{-shot} & 3\scriptsize{-shot} & 1\scriptsize{-shot} & 3\scriptsize{-shot} &
1\scriptsize{-shot} & 3\scriptsize{-shot} & 1\scriptsize{-shot} & 3\scriptsize{-shot}  \\
\hline
ENT  & S & 62.9 & 68.6 & 65.3 & 68.6 & 61.6 & 64.9 & 54.5 & 62.1 & 60.6 & 65.2 & 55.3 & 58.1 & 73.7 & 78.8 & 62.0 & 66.6 \\
MME  & S & 68.5 & 70.4 & 64.2 & 66.7 & 67.3 & 70.7 & 59.8 & 61.5 & 65.3 & 67.8 & 58.2 & 62.8 & 75.3 & 77.6 & 65.5 & 68.2 \\
KD(ROT+ENT) & S & 65.6 & 71.6 & 70.4 & 70.8 & 64.8 & 71.2 & 58.1 & 64.1 & 62.6 & 67.4 & 60.0 & 63.6 & 75.8 & 80.7 & 65.3 & 69.9 \\
KD(ROT)    & S & 63.6 & 64.0 & 66.5 & 67.0 & 60.8 & 65.0 & 57.5 & 60.9 & 63.4 & 62.6 & 58.3 & 60.3 & 74.6 & 75.6 & 63.5 & 65.1 \\
\hline \hline
ENT   & U & 52.4 & 58.2 & 59.8 & 63.2 & 53.2 & 59.8 & 49.4 & 52.7 & 54.8 & 61.5 & 45.3 & 52.3 & 70.1 & 75.2 & 55.0 & 60.4 \\
MME   & U & 59.9 & 63.1 & 64.2 & 66.6 & 60.4 & \bf{64.5} & 56.0 & 57.1 & \bf{63.2} & \bf{65.5} & 54.5 & 54.5 & 72.6 & 75.0 & 61.5 & 63.8 \\
KD(ROT+ENT)  & U & 56.1 & 64.0 & 61.9 & 66.7 & 49.8 & 58.9 & 50.8 & 57.2 & 57.4 & 61.6 & 49.0 & 55.4 & 69.2 & \bf{76.1} & 56.3 & 62.8 \\
KD(ROT)     & U & \bf{60.7} & \bf{64.8} & \bf{65.9} & \bf{67.5} & \bf{61.9} & 63.9 & \bf{57.6} & \bf{60.0} & 61.7 & 64.7 & \bf{56.1} & \bf{57.5} & \bf{73.6} & \bf{76.1} & \bf{62.5} & \bf{64.9} \\
\bottomrule
\end{tabular}
}
\end{center}
\caption{SSDA on un-curated DomainNet for ResNet. Please refer to to table \ref{tb:dn_uncurated_alexnet} for description. Again in average, our method (KD(ROT)) outperforms the baselines on this architecture.}
\label{tb:dn_uncurated_resnet}

\end{table*}

\section{Experiments}

In this section, we evaluate our method on two different domain adaptation settings using two large-scale and challenging datasets. We also conduct experiments on two different models, AlexNet and ResNet, to show the general applicability of our method. Our main focus is on the SSDA as we believe it to be a less explored area which has numerous applications.

{\bf Method names:} {\bf S+T} is a simple baseline that does only supervised training on both, source and target, domains without using any unlabeled data, {\bf ROT} is our method proposed in Eq. \ref{opt}, {\bf KD(ROT)} is our main method that performs knowledge distillation on ROT. In {\bf KD(ROT+ENT)}, we add entropy minimization loss to improve our method on the standard dataset setting.

\subsection{Datasets}
We conduct experiments on DomainNet \cite{Peng_2019_ICCV} and VisDa-17 \cite{Peng2017VisDATV}. DomainNet is a new large-scale domain adaptation dataset introduced recently. It has been used in multi-source and semi-supervised domain adaptation settings. VisDa-17 is a widely used dataset in UDA works. We primarily focus on DomainNet in this work.\\

{\bf DomainNet:}

DomainNet \cite{Peng_2019_ICCV} is a large scale domain adaptation dataset with 6 domains (Real, Clipart, Sketch, Painting, Quickdraw, and Infograph) and 345 categories. It contains about 0.6 million images. It surpasses all other previous domain adaptation datasets in terms of size and diversity.

\textbf{Standard DomainNet:} We refer the subset of DomainNet used in \cite{Saito_2019_ICCV} as standard DomainNet. This subset consists of 4 domains (Real, Clipart, Sketch, and Painting) and 126 categories. Of all possible domain pairs (source-target), 7 are chosen for evaluation. Further, two different semi-supervised settings, 1-shot and 3-shot, are created by keeping the labels for 1 and 3 samples per class while discarding the labels for the rest. We use the same dataset splits as \cite{Saito_2019_ICCV}.

\textbf{Un-curated DomainNet:} One of the reasons to choose DomainNet and particularly its subset used in semi-supervised setting is the ability to simulate true unlabeled data. We create a dataset by taking images from all 345 available categories for 4 domains in the standard DomainNet. We discard all labels and only use it as unlabeled images for target domain. We refer to this dataset as un-curated DomainNet.

For the sake of comparison, we list sizes of unlabeled images in the Table \ref{tab:domain_len}.\\
\begin{table}
    \centering
    \begin{tabular}{|c|cc|}
    \hline
        Domain & Standard & Un-curated \\\hline
        Real        & ~70k & ~175k \\
        Clipart     & ~18k & ~48k \\
        Sketch      & ~24k & ~70k \\
        Painting    & ~31k & ~75k \\
    \hline
    \end{tabular}
    \vspace{0.5em}
    \caption{Size of unlabeled target images in Standard vs. Un-curated DomainNet}
    \label{tab:domain_len}
\end{table}{}

\textbf{VisDa-17:} 

VisDa-17 \cite{Peng2017VisDATV} is a dataset for UDA. The source dataset consists of synthetic images obtained by rendering 3D models at different angles and lighting conditions. The target domain consists of images cropped from MS-COCO dataset \cite{Lin2014MicrosoftCC} using ground truth bounding boxes to only contain objects of interest. Both, source and target, domains contain 12 categories. The target dataset has ~55k images, while the source dataset has ~152k images.

\textbf{Standard VisDa-17:} We refer to the above described dataset as standard VisDa-17 dataset.

\textbf{Un-curated VisDa-17:} Similar to un-curated DomainNet, we construct un-curated VisDa-17 by adding all non-cropped training images of MS-COCO to the unlabeled target dataset. This ensures that the target domain (real images) remains intact but the unlabeled target domain contains more than just training categories. We discard labels from this dataset and only use it in an unlabeled setting.

\subsection{Semi-supervised domain adaptation}

\textbf{Standard DomainNet:} Tables \ref{tb:dn_curated_alexnet} and \ref{tb:dn_curated_resnet} compare the results of our method with state-of-the-art baselines on the standard dataset. On Alexnet architecture, our simple method of KD(ROT) outperforms the baselines. On ResNet with 3-shot, our method combined with ENT loss (KD(ROT+ENT)) outperforms the baselines.

\textbf{Un-curated DomainNet:} We compare our method with MME and ENT baselines for investigating the effect of extra categories in the unlabeled target set. For fair comparison, we rerun the official implementation of MME and ENT. Next, we run their methods on the un-curated DomainNet and report the results in Tables \ref{tb:dn_uncurated_alexnet} and \ref{tb:dn_uncurated_resnet}. In average, our method, KD(ROT), outperforms all the baselines on the un-curated dataset using both AlexNet and ResNet architecture. In Table \ref{tab:degradation_visda}, we summarize the results of Tables \ref{tb:dn_uncurated_alexnet} and \ref{tb:dn_uncurated_resnet} for understanding degradation. Note that KD(ROT) degrades the least in all cases which confirms the robustness of our method in removing the curation bias. Moreover, note that adding ENT to our method in this case degrades the result which is expected.



\begin{table}
    \centering
    \begin{tabular}{|l|cc|cc|}
    \hline
        \multirow{2}{*}{Method} & \multicolumn{2}{c}{AlexNet} & \multicolumn{2}{c|}{ResNet} \\
         & 1\scriptsize{-shot} & 3\scriptsize{-shot} & 1\scriptsize{-shot} & 3\scriptsize{-shot} \\ \hline
        ENT             & -8.2\%  & -8.1\%    & -11.3\% & -9.3\%  \\
        MME             & -6.3\%  & -5.6\%    & -6.0\% & -6.5\% \\\hline
        KD(ROT+ENT)     & -7.6\%  & -6.8\%    & -13.8\% & -10.2\% \\
        KD(ROT)         & \textbf{-3.2}\% & \textbf{-4.8}\% & \textbf{-1.6}\% & \textbf{-0.3}\% \\
    \hline
    \end{tabular}
    \vspace{0.5em}
    \caption{Degradation of the accuracy when changing the unlabeled data from standard dataset to the un-curated one for VisDa-17. Degradation percentage is relative to the standard dataset.}
    \label{tab:degradation_visda}
\end{table}

\subsection{Unsupervised domain adaptation:}

\textbf{Standard VisDa-17:} We also perform experiments for UDA where there is no labeled data in the target domain. In Table \ref{tab:st_visda}, we compare the results of our method with state-of-the-art unsupervised domain adaptation methods. We find that our method is very competitive with other methods despite being conceptually simpler. Moreover, most of the methods we compare against involve adversarial training which can be difficult to train. Apart from difficulty in training, some of the methods e.g., DTA \cite{Lee_2019_ICCV}, can involve up to 5 different loss components which means additional 5 hyper-parameters to tune. Even our most basic method, KD(ROT), is comparable to more involved methods like MCD \cite{Saito2017MaximumCD} and CDAN \cite{Long2017ConditionalAD}. We also observe significant boost in performance when our method is combined with entropy minimization (ENT) \cite{grandvalet2005semi} and virtual adversarial training (VAT) \cite{Miyato2017VirtualAT}. We show that it is possible to achieve second best performance on standard VisDa-17 benchmark with VAT and entropy minimization added to our method. VAT is a regularization method introduced in \cite{Miyato2017VirtualAT} that encourages the model to be smooth around each unlabeled data-point.


\begin{table}
    \centering
    \begin{tabular}{|l|c|}
    \hline
        Method & Mean \\\hline
        ResNet & 52.4 \\
        DANN \cite{Ganin:2015:UDA:3045118.3045244} & 57.4 \\
        MCD \cite{Saito2017MaximumCD} & 71.9 \\
        CDAN \cite{Long2017ConditionalAD} & 73.7 \\
        ADR \cite{saito2018adversarial} & 74.8 \\
        BSP+CDAN \cite{chen2019transferability} & 75.9 \\
        DTA \cite{Lee_2019_ICCV} & \bf{81.5} \\\hline
        KD(ROT) & 71.9 \\
        KD(ROT+ENT) & 74.8 \\
        KD(ROT+ENT+VAT) & 76.7 \\
    \hline
    \end{tabular}
    \vspace{0.5em}
    \caption{ResNet101 mean accuracy for various methods on VisDa-17. Our method combined with ENT and VAT is the second best.}
    \label{tab:st_visda}
\end{table}

\textbf{Un-curated VisDa-17:} We use two recent state-of-the-art methods (BSP \cite{chen2019transferability} and DTA \cite{Lee_2019_ICCV}) in UDA to understand the effect of extra categories in the unlabeled target domain. For a fair comparison, we first ran the official implementations of BSP and DTA on the standard VisDa-17 dataset to confirm the numbers. Table \ref{tab:uc_visda} shows the results for these re-runs and the actual experiments on un-curated VisDa-17.

We observe that although our method, KD(ROT), is not the best on the un-curated dataset, it degrades less compared to the other methods. The degradation in performance for our case is only 3.3\% compared to other methods where degradation ranges from ~9\%-12\%. Thus, it is reasonable to conclude that the huge boost is performance by recent UDA methods is tied to the assumption of curated unlabeled target samples. This makes our method a very simple and effective baseline for future comparison.

\begin{table}
    \centering
    \begin{tabular}{|l|ccc|}
    \hline
         Method & Standard & Un-curated & Degradation\\\hline
         BSP+CDAN & 75.9 & 68.6 & -9.6\% \\
         cDTA+fDTA & 77.4 & 68.5 & -11.5\% \\
         DTA & \textbf{81.1}& \textbf{71.8} & -11.5\% \\\hline
         KD(ROT) & 71.9 & 69.5 & \textbf{-3.3\%} \\
    \hline
    \end{tabular}
    \vspace{0.5em}
    \caption{ResNet101 mean accuracy for various methods on un-curated VisDa-17. For standard dataset, we report the results of rerunning the baselines for fair comparison. cDTA+fDTA refers to DTA\cite{Lee_2019_ICCV} without VAT introduced in \cite{Lee_2019_ICCV}. The degradation percentage is relative to the standard dataset. Our method leads to least degradation when we go from standard dataset to the un-curated one.}
    \label{tab:uc_visda}
\end{table}

\subsection{Jigsaw vs. Rotation:} 

Since \cite{Carlucci_2019_CVPR} uses Jigsaw solver in domain generalization setting. Here, we study using Jigsaw instead of RotNet in our SSDA setting. We tried $\lambda_{ssl} \in \{0.7, 1.0\}$ and $\lambda_{ent} \in \{0.01, 0.1\}$ for real to sketch pair and picked the best combination. These are the parameters used in \cite{Carlucci_2019_CVPR}. We do not do knowledge distillation for these experiments. We list our results in Table \ref{tab:jig_rot}. We found that Jigsaw results are close to the S+T baseline. Note that Jigsaw is shown to be more effective than RotNet in \cite{Carlucci_2019_CVPR}, but it is significantly worse than RotNet in our SSDA setting. We empirically conclude that Jigsaw does not generalize well in the case of single source domain.

\begin{table}
    \centering
    \begin{tabular}{|c|cc|}
    \hline
        Method & AlexNet & ResNet \\\hline
        S+T & 43.4 & 60.0 \\
        ROT+ENT & 48.6 & 68.5 \\
        JIG+ENT & 44.1 & 59.5 \\
    \hline
    \end{tabular}
    \vspace{0.5em}
    \caption{Mean accuracies for 3-shot SSDA. We don't do knowledge distillation for these experiments.}
    \label{tab:jig_rot}
\end{table}

\subsection{Implementation details}
Our code is implemented in PyTorch and closely follows the implementation of \cite{Saito_2019_ICCV}.
Because the focus of this work is on obtaining baselines using methods that are easy to train, we refrain
from extensive hyper-parameter tuning. In all experiments, we use the

\textbf{Semi-supervised domain adaptation:}
We use AlexNet\cite{krizhevsky2012imagenet} and ResNet34 \cite{he2016deep} pre-trained on ImageNet
in all of our experiments. The architectures of feature extractor and supervised classification
head are kept the same as in \cite{Saito_2019_ICCV} for fair comparison.
For self-supervised classification head, we use a single fully-connected layer for
AlexNet, while two fully-connected layers with a ReLU activation between them is used for ResNet34. We use the
same learning rate annealing schedule as in \cite{Ganin:2015:UDA:3045118.3045244}. The model is optimized using
SGD with a momentum of 0.9 and weight decay of $0.0005$. The initial learning rate for feature extractor is 0.001 while for both classification heads it is 0.01.

We use $\lambda_{ent} = 0.01$ for AlexNet and
$\lambda_{ent} = 0.1$ for ResNet34. We tried $\lambda_{ent} \in \{0.1, 0.01, 0.05\}$
values on Real to Sketch pair for 3-shot setting. We use $\lambda_{ssl} = 1, \lambda_s = \lambda_t = 1$ for all experiments. We selected these parameters without any tuning. The training is run for 30k iterations and the checkpoint with best validation accuracy is used for testing. 

\textbf{Unsupervised Domain Adaptation:} For a fair comparison with other works, we only use ResNet101 \cite{he2016deep}. Apart from weights for losses, all other hyper-parameters are the same as above. We search for $\lambda_s \in \{0.5, 1.0\}$ and $\lambda_{ent} \in \{0.05, 0.01, 0.1\}$. We use $\lambda_{s}=0.5, \lambda_{ent} = 0.01, \lambda_{vat} = 0.01$ for all our experiments when the corresponding losses are used. We use $\lambda_{vat} = 0.01$, without any tuning. Also, the parameters for the VAT \cite{Miyato2017VirtualAT} loss are the same as those in the original work.

\textbf{Knowledge distillation:} We start with an ImageNet pre-trained student and run the training for 10 epochs while dropping learning rate by a factor of 0.1 every 3 epochs. We intentionally keep the number of epochs for distillation small to reduce computational time and keep the experiments simple. 

\section{Conclusion}
We introduced a simple method for domain adaptation that utilizes a self-supervised learning task (RotNet) on the unlabeled target data along with supervised learning on the source domain. Even though simple, either our method outperforms or is on par with state-of-the-art methods on two challenging benchmarks. Moreover, we point out that unlabeled target data in those two popular benchmarks has a selection bias that may not exist in a real application. We show that by removing this bias, most methods degrade in performance while ours is relatively robust. We believe our simple method and also un-curated setting can have a good impact on evaluating the future domain adaptation research.

\vspace{1in}
\section{Appendix}

We perform more experiments to study the effect of different parts of our method. We study Ablation on VisDa-17 on Table \ref{tab:ablation}, the effect of the choice of dataset for distillation on VisDa-17 on Table \ref{tab:st_vd_st_vs_uc_dist}, the effect of the choice of dataset for distillation on VisDa-17 on Table \ref{tab:uc_vd_st_vs_uc_dist}, training RotNet on both domains for DomainNet on Table \ref{tab:source_ssl}, ensembling for DomainNet on table \ref{tab:multi_dist}, the effect of pretraining for DomainNet on table \ref{tab:multi_pretrain_dist}, the effect of distillation for AlexNet on DomainNet on Table \ref{tb:dn_st_alexnet_dist}, and the effect of distillation for ResNet on DomainNet on Table \ref{tb:dn_st_resnet_dist}. Please read the captions of the tables for the description.

\begin{table}[thbp]
    \centering
    \begin{tabular}{|l|c|c|c|}
    \hline
        Method & Standard & Un-curated & Degrad. \\\hline
        Source only     & 57.6  & 57.6 & 0.0\%  \\ 
        KD(Source only) & 60.9  & 59.8 & -1.8\% \\
        ENT             & 69.4  & 65.1 & -6.2\% \\ 
        VAT             & 65.7  & 64.1 & -2.4\% \\ 
        ENT+VAT         & 69.8  & 66.9 & -4.2\% \\ 
        ROT             & 69.2  & 67.9 & -1.9\% \\ 
        ROT+ENT         & 73.1  & 67.1 & -8.2\% \\ 
        ROT+VAT         & 71.5  & 69.2 & -3.2\% \\ 
        ROT+ENT+VAT     & 74.9  & 67.1 & -10.4\% \\ 
        KD(ROT+VAT)         & 73.7  & {\bf 70.8} & -3.9\% \\ 
        KD(ROT+ENT+VAT)     & {\bf 76.7}  & 68.7 & -10.4\% \\ 
    \hline
    \end{tabular}
    \vspace{0.5em}
    \caption{{\bf Ablation study on VisDa-17 (ResNet101):} We evaluate different combinations of methods. ``Source only'' is the baseline model that is supervised on the source domain only which is equivalent to the first row of Table 7 in the main paper. Note that the accuracy is slightly higher since we run this baseline ourselves rather than copying it from [8]. The second row shows that knowledge distillation alone improves this baseline with a margin that is smaller than our main method. Also, in the last column, we also show the degradation of the accuracy when we change the unlabeled dataset from standard to un-curated. As expected, adding ENT hurts in the case of un-curated data. ROT and KD(Source only) have the least amount of degradation. KD(ROT+VAT) achieves the best accuracy for un-curated data.
    }
    \label{tab:ablation}
\end{table}





\begin{table}[tbph]
    \centering
    \begin{tabular}{|l|c|c|c|}
    \hline
        Method & no KD. & KD Std. & KD UnC. \\\hline
        ROT             & 69.2 & 71.9 & 70.7 \\
        ROT+ENT         & 73.1 & 74.8 & 74.1 \\
        ROT+VAT         & 71.5 & 73.7 & 73.4 \\
        ROT+ENT+VAT     & 74.9 & 76.7 & 75.9 \\
    \hline
    \end{tabular}
    \vspace{0.5em}
    \caption{{\bf Effect of the choice of dataset for distillation on VisDa-17 (ResNet101):} We train a model with no distillation for different combinations of methods on standard unlabeled data (column ``no KD''). Then, we show the results of distillation using standard and un-curated data separately. We do not observe a large degradation in accuracy when we change the dataset of distillation from standard to un-curated. This shows that distillation is robust to curation bias.}
    \label{tab:st_vd_st_vs_uc_dist}
\end{table}

\begin{table}[tbhp]
    \centering
    \begin{tabular}{|l|ccc|}
    \hline
        Method & no KD. & KD Std. & KD UnC. \\\hline
        ROT             & 67.9 & 70.7 & 69.5 \\
        ROT+ENT         & 67.1 & 69.1 & 69.9 \\
        ROT+VAT         & 69.2 & 72.1 & 70.8 \\
        ROT+ENT+VAT     & 67.1 & 68.7 & 68.7 \\
    \hline
    \end{tabular}
    \vspace{0.5em}
    \caption{{\bf Effect of the choice of dataset for distillation on VisDa-17 (ResNet101):} This tabel is similar to Table \ref{tab:st_vd_st_vs_uc_dist} except that we use un-curated data for the methods before distillation. Again, we observe that knowledge distillation is robust to the dataset curation bias.}
    \label{tab:uc_vd_st_vs_uc_dist}
\end{table}

\begin{table*}[h]
\begin{center}
\renewcommand{\arraystretch}{1.1}
\begin{tabular}{|c|l|ccccccc|c|}
\hline
    Model & Method & R to C & R to P & P to C & C to S & S to P & R to S & P to R & Mean \\
    \hline
    \multirow{2}{*}{AlexNet}
    & ROT(t)+ENT      & 52.9 & 49.6 & 51.1 & 41.2 & 44.3 & 40.7 & 60.1 & 48.6 \\
    & ROT(s+t) + ENT  & 53.1 & 48.0 & 50.9 & 41.7 & 43.0 & 41.6 & 60.6 & 48.4 \\
    \hline
    
    \multirow{2}{*}{ResNet}
    & ROT(t)+ENT      & 70.4 & 69.4 & 69.9 & 62.8 & 65.8 & 62.3 & 79.1 & 68.5 \\
    & ROT(s+t) + ENT  & 70.9 & 70.0 & 70.4 & 61.2 & 65.0 & 64.0 & 77.7 & 68.5 \\
\hline
\end{tabular}
\vspace{0.5em}
\caption{{\bf Training RotNet on both domains for DomainNet:} We use 3-shot standard DomainNet for these experiments. ROT(t) is our main method that only uses the target dataset for the RotNet loss. We show that ROT(s+t), uses the source dataset as well for the RotNet loss, does not improve the final model.}
\label{tab:source_ssl}
\end{center}
\end{table*}

\vspace{-.2in}

\begin{table*}[h]
\begin{center}
\begin{tabular}{|l|ccccccc|c|c|}
\hline
    type & P to R & S to P & C to S & P to C & R to P & R to C & R to S & Mean & \\\hline
    1. $\text{ROT}_1$                                 & 60.4 & 44.3 & 41.5 & 49.7 & 49.9 & 51.6 & 40.1 & 48.2 & 1 \\
    2. $\text{ROT}_2$                                 & 60.3 & 44.6 & 41.6 & 49.8 & 49.9 & 51.8 & 39.9 & 48.3 & 2 \\
    3. $\text{MME}_1$                                 & 61.1 & 44.4 & 40.5 & 50.7 & 50.1 & 54.7 & 38.7 & 48.6 & 3 \\
    4. $\text{MME}_2$                                 & 61.5 & 44.8 & 41.2 & 50.6 & 50.2 & 55.2 & 38.6 & 48.9 & 4 \\
    5. $\text{ens}(\text{ROT}_1, \text{ROT}_2)$                  & 61.2 & 45.5 & 42.6 & 50.9 & 50.7 & 52.7 & 41.1 & 49.2 & 5 \\
    6. $\text{ens}(\text{MME}_1, \text{MME}_2)$                 & 62.4 & 45.7 & 42.3 & 51.8 & 51.2 & 56.0 & 39.7 & 49.9 & 6 \\
    7. $\text{ens}(\text{MME}_1, \text{ROT}_1)$                 & 62.6 & 47.3 & 43.0 & 52.5 & 51.8 & 56.0 & 41.0 & 50.6 & 7 \\
    8. $\text{KD}(\text{ens}(\text{ROT}_1, \text{ROT}_2))$             & 62.8 & 46.9 & 43.8 & 52.3 & 51.9 & 54.1 & 42.4 & 50.6 & 8 \\
    9. $\text{KD}(\text{ens}(\text{MME}_1, \text{MME}_2))$             & 63.6 & 46.5 & 43.3 & 52.4 & 52.4 & 56.9 & 40.5 & 50.8 & 9 \\
    10. $\text{KD}(\text{ens}(\text{MME}_1, \text{ROT}_1))$             & 64.0 & 48.6 & 44.2 & 53.7 & 53.1 & 57.0 & 42.0 & 51.8 & 10 \\
\hline
\end{tabular}
\end{center}
\caption{{\bf Ensembling results for DomainNet:} We use 3-shot standard DomainNet and AlexNet for these experiments. ``$\text{ens}$'' means ensembling two different models. We are interested in studying the effect of using multiple methods. Using a fixed initialization, we run our method twice to get ($\text{ROT}_1$ and $\text{ROT}_2$) and MME method twice to get ($\text{MME}_1$ and $\text{MME}_2$). We show that we get the best accuracy by combining one of ROT and one of MME models. This accuracy is better than combining two models of the same kind. This is a simple way of combining ROT and MME compared to training them jointly.
}
\label{tab:multi_dist}
\end{table*}


\begin{table*}[h]
\begin{center}
\begin{tabular}{|l|ccccccc|c|c|}
\hline
    Method & P to R & S to P & C to S & P to C & R to P & R to C & R to S & Mean & \\\hline
    1. $\text{ROT}_{sup1}$                                 & 60.4 & 44.3 & 41.5 & 49.7 & 49.9 & 51.6 & 40.1 & 48.2 & 1 \\
    2. $\text{ROT}_{sup2}$                                 & 60.3 & 45.2 & 41.9 & 49.6 & 49.8 & 51.4 & 40.5 & 48.3 & 2 \\
    3. $\text{ROT}_{sup+rot}$                         & 59.2 & 44.0 & 42.6 & 49.9 & 49.7 & 52.4 & 41.3 & 48.4 & 3 \\
    4. $\text{ens}(\text{ROT}_{sup1}, \text{ROT}_{sup2})$                  & 62.3 & 47.0 & 43.7 & 52.1 & 51.9 & 53.5 & 42.3 & 50.4 & 4 \\
    5. $\text{ens}(\text{ROT}_{sup1}, \text{ROT}_{sup+rot})$          & 61.9 & 46.8 & 44.4 & 52.8 & 52.0 & 54.3 & 43.2 & 50.8 & 5 \\
    6. $\text{KD}(\text{ens}(\text{ROT}_{sup1}, \text{ROT}_{sup+rot}))$     & 63.4 & 48.3 & 45.5 & 54.1 & 53.1 & 55.6 & 44.2 & 52.0 & 6 \\
\hline
\end{tabular}
\end{center}
\caption{{\bf The effect of pretraining for DomainNet:} We use 3-shot standard DomainNet and AlexNet for these experiments. We initialize our method with three different pretrained models ($\text{ROT}_{sup1}$, $\text{ROT}_{sup2}$, and $\text{ROT}_{sup+rot}$). The first two are regular supervised pretrainings on ImageNet and the last one uses both, supervised loss and RotNet self-supervised loss, on ImageNet. We show that adding RotNet to the pretraining helps slightly to generalize better. We also show that using this model as one of the teachers in ensembling and distillation helps by almost 4 points.
}
\label{tab:multi_pretrain_dist}
\end{table*}

\begin{table*}[t]
\begin{center}
\scalebox{0.85}
{
\renewcommand{\arraystretch}{1.1}
\begin{tabular}{lc|cccccccccccccc|cc}
\toprule

\multirow{2}{*}{Method} & \multirow{2}{*}{\hspace{-.2in}Data} &
\multicolumn{2}{c}{R to C} & \multicolumn{2}{c}{R to P} & \multicolumn{2}{c}{P to C} &
\multicolumn{2}{c}{C to S} & \multicolumn{2}{c}{S to P} & \multicolumn{2}{c}{R to S} &
\multicolumn{2}{c}{P to R} & \multicolumn{2}{|c}{Mean} \\

& &
1\scriptsize{-shot} & 3\scriptsize{-shot} & 1\scriptsize{-shot} & 3\scriptsize{-shot} & 1\scriptsize{-shot} & 3\scriptsize{-shot} &
1\scriptsize{-shot} & 3\scriptsize{-shot} & 1\scriptsize{-shot} & 3\scriptsize{-shot} & 1\scriptsize{-shot} & 3\scriptsize{-shot} &
1\scriptsize{-shot} & 3\scriptsize{-shot} & 1\scriptsize{-shot} & 3\scriptsize{-shot}  \\
\hline

ROT         & S & 47.7 & 51.9 & 48.3 & 49.5 & 45.4 & 50.3 & 38.6 & 41.3 & 42.0 & 44.2 & 38.5 & 41.5 & 56.7 & 59.5 & 45.3 & 48.3 \\
KD(ROT)     & S & 49.3 & 54.5 & 49.9 & 51.3 & 48.3 & 52.9 & 39.7 & 43.2 & 44.3 & 46.3 & 40.0 & 43.2 & 58.7 & 61.5 & 47.2 & 50.4 \\
ROT+ENT     & S & 49.3 & 52.9 & 48.9 & 49.6 & 45.6 & 51.1 & 36.8 & 41.2 & 36.7 & 44.3 & 37.5 & 40.7 & 58.1 & 60.1 & 44.7 & 48.6 \\
KD(ROT+ENT) & S & 51.0 & 54.7 & 50.5 & 50.9 & 47.8 & 53.0 & 37.7 & 42.3 & 38.1 & 46.3 & 38.0 & 41.8 & 60.4 & 62.1 & 46.2 & 50.2 \\
\hline
ROT         & U & 45.8 & 49.8 & 47.4 & 48.2 & 44.3 & 47.8 & 36.7 & 40.8 & 40.8 & 42.3 & 37.1 & 39.3 & 56.6 & 59.3 & 44.1 & 46.8 \\
KD(ROT)     & U & 47.7 & 51.3 & 48.9 & 48.3 & 46.5 & 50.2 & 37.4 & 41.7 & 42.7 & 43.8 & 38.1 & 40.5 & 58.5 & 60.3 & 45.7 & 48.0 \\
ROT+ENT     & U & 45.2 & 50.0 & 46.3 & 48.7 & 42.1 & 46.8 & 34.3 & 38.5 & 34.1 & 40.2 & 33.3 & 38.9 & 56.4 & 59.2 & 41.7 & 46.0 \\
KD(ROT+ENT) & U & 46.4 & 48.9 & 47.2 & 50.0 & 43.7 & 49.0 & 34.3 & 39.2 & 35.8 & 41.8 & 33.3 & 38.1 & 57.9 & 60.5 & 42.7 & 46.8 \\

\bottomrule
\end{tabular}
}
\end{center}
\caption{{\bf Effect of distillation for AlexNet on DomainNet:} We study the effect of distillation on different variations of the method. The conclusion is that distillation helps the model in all experiments. The first section uses standard data while the second one uses un-curated data.}
\label{tb:dn_st_alexnet_dist}

\end{table*}


\begin{table*}[!t]
\begin{center}
\scalebox{0.85}
{
\renewcommand{\arraystretch}{1.2}
\begin{tabular}{lc|cccccccccccccc|cc}
\toprule
\multirow{2}{*}{Method} & \multirow{2}{*}{\hspace{-.2in}Data} &
\multicolumn{2}{c}{R to C} & \multicolumn{2}{c}{R to P} & \multicolumn{2}{c}{P to C} &
\multicolumn{2}{c}{C to S} & \multicolumn{2}{c}{S to P} & \multicolumn{2}{c}{R to S} &
\multicolumn{2}{c}{P to R} & \multicolumn{2}{|c}{Mean} \\

& &
1\scriptsize{-shot} & 3\scriptsize{-shot} & 1\scriptsize{-shot} & 3\scriptsize{-shot} & 1\scriptsize{-shot} & 3\scriptsize{-shot} &
1\scriptsize{-shot} & 3\scriptsize{-shot} & 1\scriptsize{-shot} & 3\scriptsize{-shot} & 1\scriptsize{-shot} & 3\scriptsize{-shot} &
1\scriptsize{-shot} & 3\scriptsize{-shot} & 1\scriptsize{-shot} & 3\scriptsize{-shot}  \\
\hline

ROT         & S & 60.7 & 62.2 & 64.0 & 64.8 & 58.2 & 62.2 & 55.3 & 57.9 & 60.0 & 60.8 & 55.5 & 56.7 & 72.3 & 73.2 & 60.9 & 62.6 \\
KD(ROT)     & S & 63.6 & 64.0 & 66.5 & 67.0 & 60.8 & 65.0 & 57.5 & 60.9 & 63.4 & 62.6 & 58.3 & 60.3 & 74.6 & 75.6 & 63.5 & 65.1 \\
ROT+ENT     & S & 64.3 & 70.3 & 68.9 & 69.4 & 64.6 & 69.9 & 56.6 & 62.8 & 61.2 & 65.8 & 58.2 & 62.3 & 74.5 & 79.1 & 64.1 & 68.5 \\
KD(ROT+ENT) & S & 65.6 & 71.6 & 70.4 & 70.8 & 64.8 & 71.2 & 58.1 & 64.1 & 62.6 & 67.4 & 60.0 & 63.6 & 75.8 & 80.7 & 65.3 & 69.9 \\
\hline
ROT         & U & 58.7 & 62.4 & 63.9 & 64.9 & 59.0 & 61.2 & 55.0 & 58.1 & 59.1 & 61.7 & 54.9 & 56.9 & 71.8 & 74.4 & 60.4 & 62.8 \\
KD(ROT)     & U & 60.7 & 64.8 & 65.9 & 67.5 & 61.9 & 63.9 & 57.6 & 60.0 & 61.7 & 64.7 & 56.1 & 57.5 & 73.6 & 76.1 & 62.5 & 64.9 \\
ROT+ENT     & U & 54.4 & 61.7 & 58.9 & 64.9 & 48.0 & 57.1 & 48.9 & 55.5 & 55.1 & 60.9 & 46.7 & 53.3 & 68.1 & 74.6 & 54.3 & 61.1 \\
KD(ROT+ENT) & U & 56.1 & 64.0 & 61.9 & 66.7 & 49.8 & 58.9 & 50.8 & 57.2 & 57.4 & 61.6 & 49.0 & 55.4 & 69.2 & 76.1 & 56.3 & 62.8 \\
\bottomrule
\end{tabular}
}
\end{center}
\caption{{\bf Effect of distillation for ResNet on DomainNet:} Please refer to the caption of Table \ref{tb:dn_st_alexnet_dist} for the description. Again, we see that distillation always improves the model performance.}
\vspace{10in}
\label{tb:dn_st_resnet_dist}

\end{table*}
\end{document}